\documentclass[letterpaper, 10 pt, journal, twoside]{IEEEtran}  

\usepackage{tikz}
\usepackage{graphics} 
\usepackage{epsfig} 
\usepackage{amsmath} 
\usepackage{amssymb}  
\usepackage[font=footnotesize, caption=false]{subfig}
\usepackage{mathtools}

\usepackage{tikzscale}
\usepackage{algorithm}
\usepackage{subfig}
\usepackage[noend]{algpseudocode}
\usepackage[flushleft]{threeparttable}

\algnewcommand{\LeftComment}[1]{\Statex \textit{#1}}

\usepackage{color}
\definecolor{blue(pigment)}{rgb}{0.2, 0.2, 0.6}
\definecolor{asparagus}{rgb}{0.53, 0.66, 0.42}

\DeclareMathOperator*{\argmin}{argmin}
\DeclareMathOperator*{\argmax}{argmax}
\DeclareMathOperator{\CS}{\mathbb{CS}}
\DeclareMathOperator{\CSp}{\mathbb{CS}^\prime}
\DeclareMathOperator{\Cam}{\mathbb{C}}
\DeclareMathOperator{\Pat}{\mathbb{P}}
\DeclareMathOperator{\Tim}{\mathbb{T}}

\usetikzlibrary{snakes,arrows,shapes}

\newcommand{\multrow}[1]{\begin{tabular}{@{}l@{}} #1 \end{tabular}}

\usepackage[breaklinks=true,colorlinks,bookmarks=false]{hyperref}

\title{\LARGE \bf
Multi-camera calibration with pattern rigs, including for non-overlapping cameras: CALICO
}

\author{Amy Tabb$^{1}$, Henry Medeiros$^{2}$, Mitchell J. Feldmann$^{3}$, and Thiago T. Santos$^{4}$ 
\thanks{$^{1}$USDA-ARS-AFRS
        {\tt\small amy.tabb@usda.gov}}%
\thanks{$^{2}$University of Florida
        {\tt\small hmedeiros@ufl.edu}}%
\thanks{$^{3}$University of California-Davis
	{\tt\small mjfeldmann@ucdavis.edu}}%
\thanks{$^{4}$Embrapa Digital Agriculture
	{\tt\small thiago.santos@embrapa.br}}%
}

\begin{document}

\maketitle
\thispagestyle{empty}
\pagestyle{empty}

\begin{abstract}

This paper describes CALICO, a method for multi-camera calibration suitable for challenging contexts: stationary and mobile multi-camera systems, cameras without overlapping fields of view, and non-synchronized cameras. Recent approaches are roughly divided into infrastructure- and pattern-based. Infrastructure-based approaches use the scene's features to calibrate, while pattern-based approaches use calibration patterns. Infrastructure-based approaches are not suitable for stationary camera systems, and pattern-based approaches may constrain camera placement because shared fields of view or extremely large patterns are required.

CALICO is a pattern-based approach, where the multi-calibration problem is formulated using rigidity constraints between patterns and cameras. We use a {\it pattern rig}: several patterns rigidly attached to each other or some structure. We express the calibration problem as that of algebraic and reprojection error minimization problems. Simulated and real experiments demonstrate the method in a variety of settings. CALICO compared favorably to Kalibr. Mean reconstruction accuracy error was $\le 0.71$ mm for real camera rigs, and $\le 1.11$ for simulated camera rigs. Code and data releases are available at \cite{tabb_amy_2019_3520866} and \url{https://github.com/amy-tabb/calico}.\footnote{Mention of trade names or commercial products in this publication is solely for the purpose of providing specific information and does not imply recommendation or endorsement by the U.S. Department of Agriculture.  USDA is an equal opportunity provider and employer.} 
\end{abstract}

\section{Introduction}
\label{sec:init}

Multi-camera calibration is necessary for a variety of activities, from human activity detection and recognition to reconstruction tasks. Internal parameters can typically be computed by waving a calibration target in front of cameras, and then using Zhang's algorithm \cite{zhang_flexible_2000} to calibrate individual cameras.  Determining cameras' external parameters, or the relationships between cameras in a multi-camera system, may be a more difficult problem as methods for computing these relationships depend strongly on characteristics of the hardware and the arrangement of the cameras.  For instance, the cameras' shared field of view, level of synchronization, and mobility strongly influence the ease of multi-camera calibration or choice of methods available for calibration.

This work describes CALICO,\footnote{CALICO: CALibratIon of asynchronous Camera netwOrks. Acronym generated by \cite{cook_acronym:_2019} for a prior version of this work.} a method for multi-camera calibration.  We assume that cameras can be triggered to acquire images of a stationary calibration rig, or the cameras are synchronized. Our method uses calibration objects based on one or more planar patterns \cite{garrido-jurado_automatic_2014} that allow for partial pattern detection, allows non-overlapping fields of view, and allows for mobile or stationary multi-camera systems. Updated datasets and code described in this paper are at provided at \cite{tabb_amy_2019_3520866} and \url{https://doi.org/10.5281/zenodo.3520865}. Prior versions of this work are available on arXiv \cite{tabb2019calibration}.  

\newcommand{\figwidth}{.45}
\newcommand{\boxIt}{{\it{box}}}
\newcommand{\robot}{{\it{robot}}}
\newcommand{\wbs}{{\it{wide baseline stereo}}}
\newcommand{\stereo}{{\it{stereo}}}

\begin{figure}[t]
	\centering
	\subfloat[Datasets Net-1, Net-2; \boxIt \ type]
	{
		\includegraphics[width=\figwidth \linewidth]{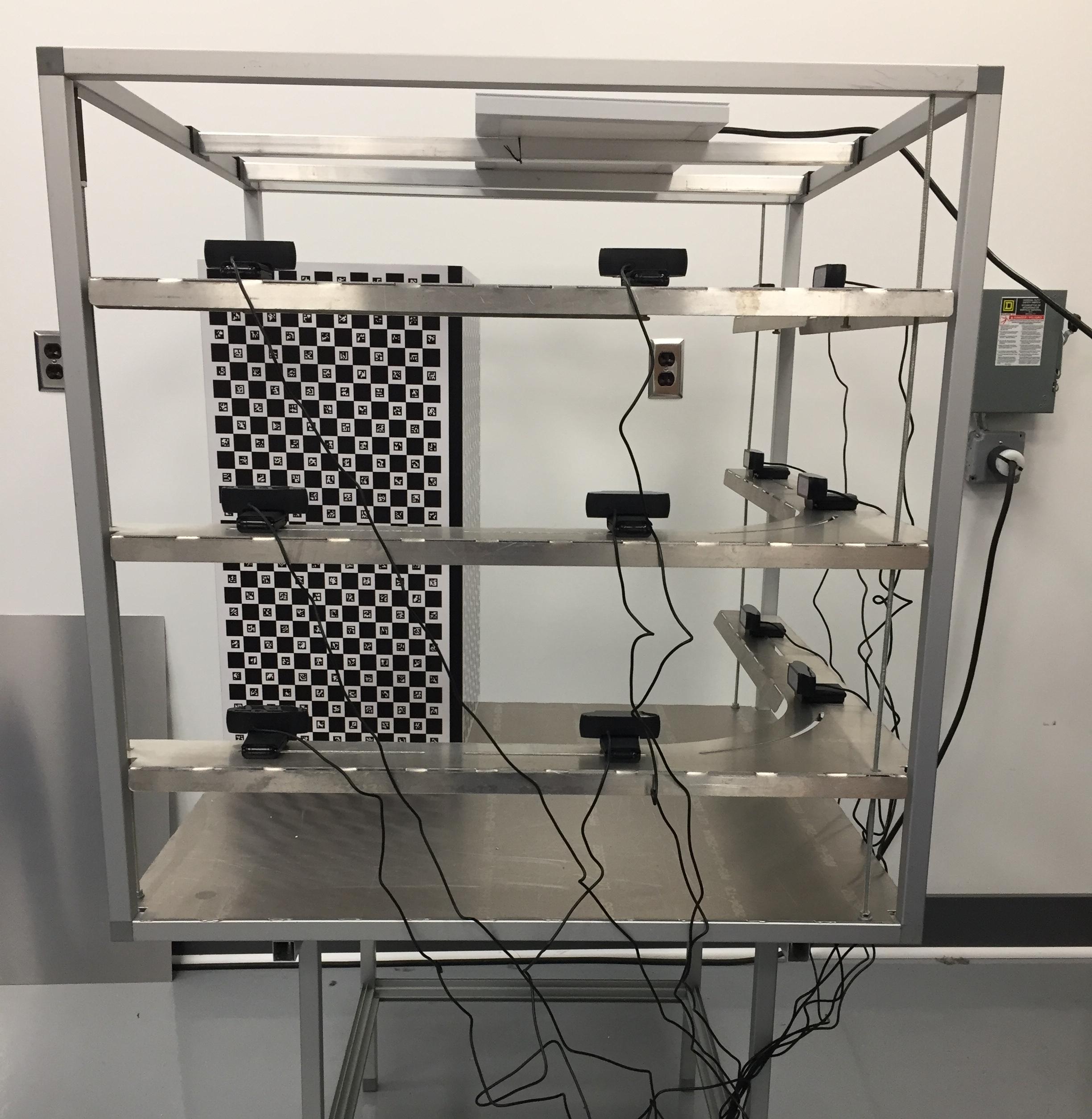} 
	} \label{sf:camera-network}
	
	\subfloat[Dataset Mult-1; \robot \ type]
	{
		\includegraphics[width=\figwidth \linewidth]{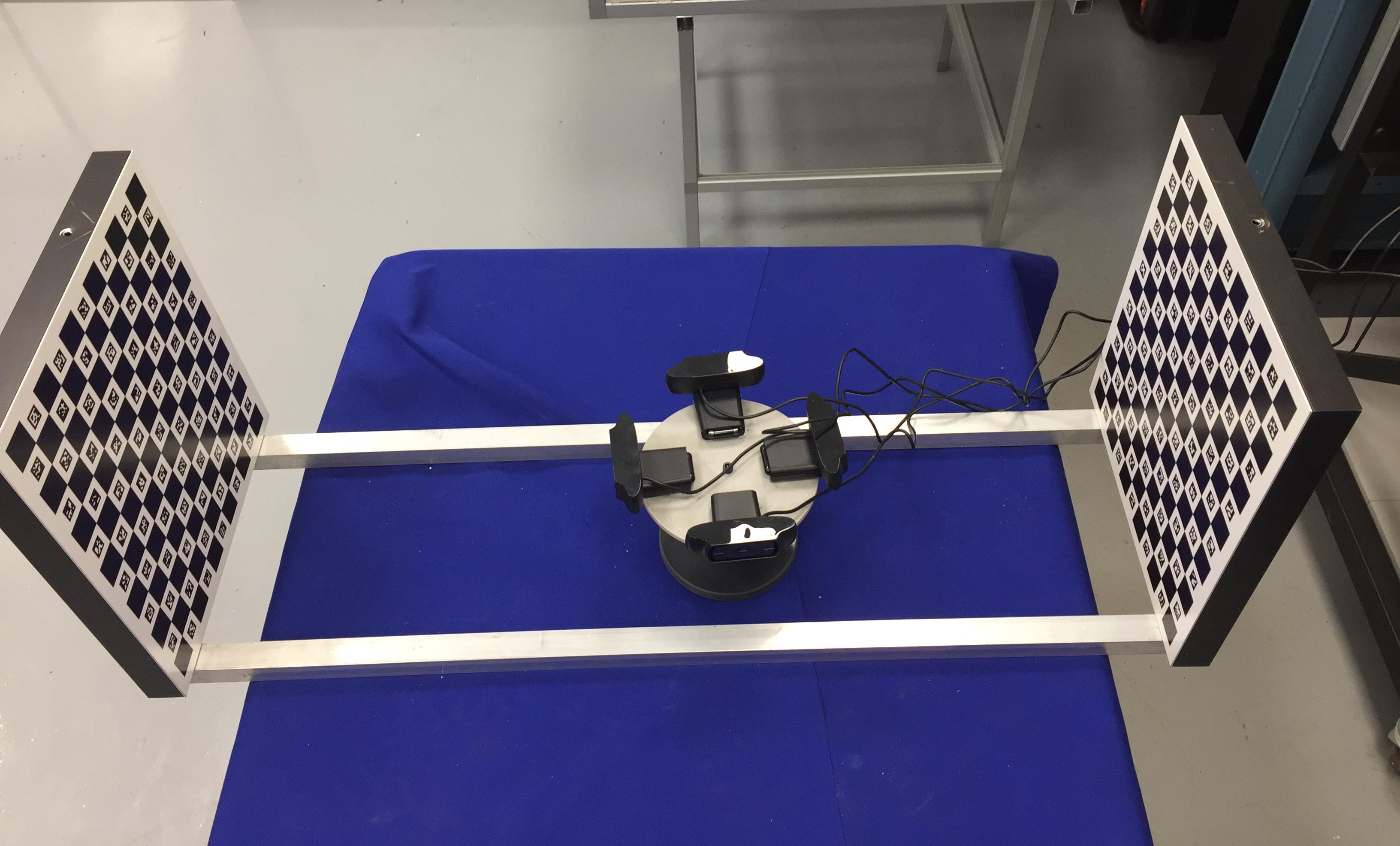} 
	} \label{sf:mult1}
	\subfloat[Dataset Mult-2; \robot \ type]
	{
		\includegraphics[width=\figwidth \linewidth]{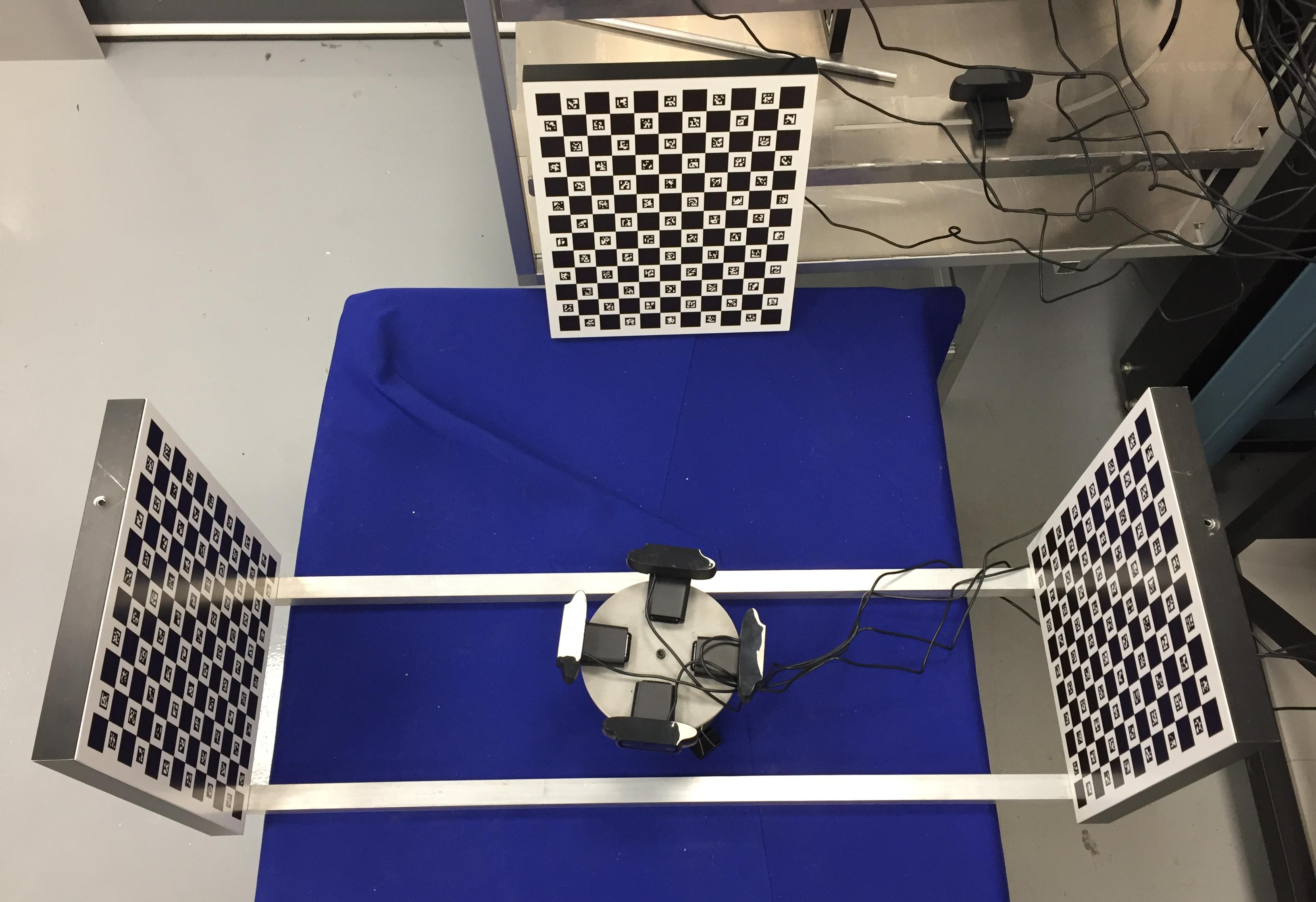}
	} \label{sf:mult2}
	\caption{{\bf Best viewed in color.} Cameras and patterns for two types of real experiments, \boxIt \ and \robot. The top row, (a) is an arrangement of cameras mounted on the periphery of a box where cameras point towards the box's interior; the bottom row, (b) and (c) are multi-camera camera rigs where cameras point away from each other. In the top figure, there are 12 cameras.  The pattern rig is moved within the workspace, and cameras acquire an image for every move. The bottom row shows a multicamera system, in (b) only two cameras are used while in (c) all four cameras are used.  In the multicamera system, the multicamera system is moved and the patterns are stationary. Each of these experiment types are challenging to calibrate using existing methods.  In the \boxIt \ type, cameras share a common field of view but cameras on one side of the box may not be able to view the same planar pattern as cameras on another side.  In the \robot \ configuration, there may be no shared field of view between cameras. CALICO approximately solves the set of constraints resulting from the camera, pattern, and time relationships in such datasets.}
	\label{fig:experimental_hardware}
\end{figure}

\begin{figure*}
	\centering
	\subfloat[Sim-2 dataset, \boxIt \ type]
	{
		\includegraphics[width=0.38\linewidth]{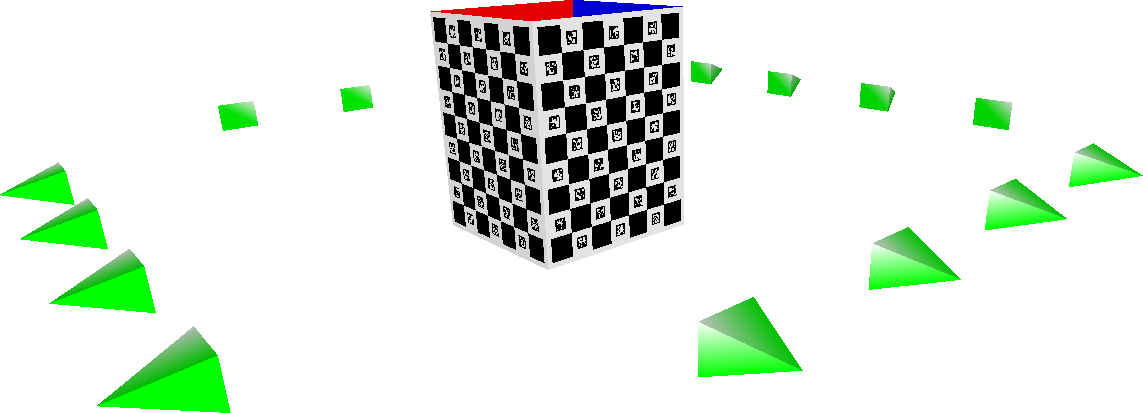} 
	} 
	\subfloat[Sim-5-april dataset, \robot \ type]
	{  
		\includegraphics[width = 0.6\linewidth]{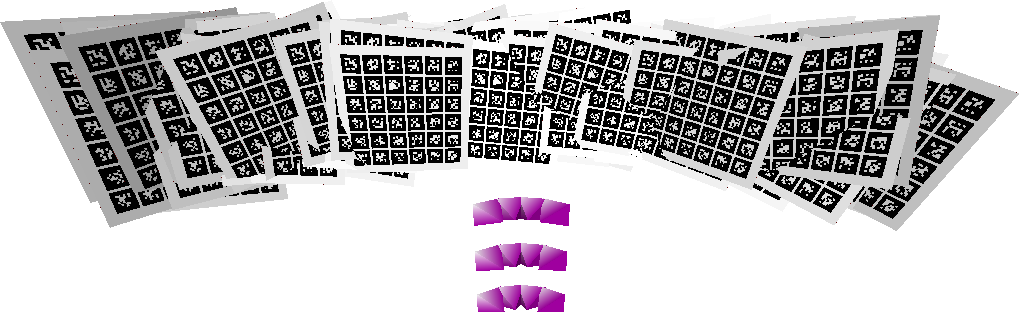} 
		
	}
	
	\subfloat[Sim-7-april dataset, \robot \ type]
	{  
		\includegraphics[width = 0.30\linewidth]{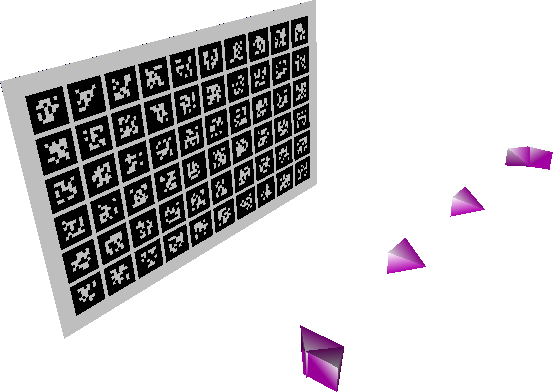} 
	}
	\subfloat[Sim-8-april dataset, \wbs \ type]
	{  
		\includegraphics[width = 0.6\linewidth]{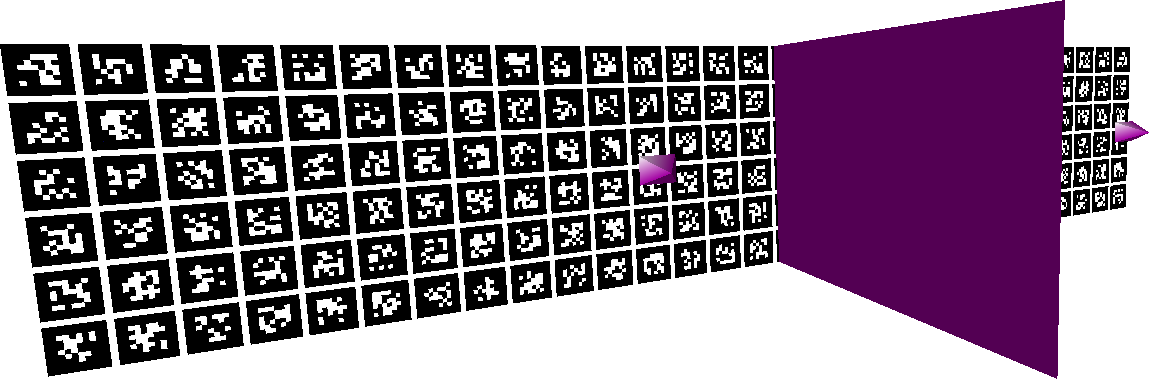} 
	}
	\caption{{\bf Examples of simulated datasets.} Arrangements of cameras for four of nine simulated datasets. (a) shows cameras as pyramids arranged on the edges of a square, and a pattern rig of four charuco patterns at one time step, a \boxIt \ type experiment. (b) shows cameras as pyramids and the placement of April Tag calibration patterns over all time steps, a \robot \ type experiment. (c) has cameras from dataset Sim-7-april and one April Tag calibration pattern, another \robot \ type experiment. (d) has the two cameras from Sim-8-april and a very large and long April Tag calibration pattern; there is a barrier so the cameras each view mutually exclusive portions of the pattern.}
	\label{fig:simulatedDatasets}
\end{figure*}

Given these preliminaries, our contributions to the state-of-the-art consist of:
\begin{enumerate} 
	\item{A formulation of the multi-camera calibration problem as a set of rigidity constraints imposed by the transformations between cameras and calibration patterns.}
	\item{The multi-camera calibration problem is solved efficiently by iteratively solving for variables using closed-form methods, minimizing algebraic error over the the set of rigidity constraints, and minimization of reprojection errors.}
\end{enumerate}

\section{Related Work}

\newcommand{\etal}{\emph{et al.}}
\newcommand{\etals}{\emph{et al. }}
\newcommand{\mishra}{Mishra \etals 2022 \cite{mishra_look_2022}}
\newcommand{\li}{Li \etals 2013 \cite{li_bo_multiple-camera_2013}}
\newcommand{\liucaliber}{Liu \etals 2016 \cite{liu_caliber:_2016}}
\newcommand{\lin}{Lin \etals 2020 \cite{lin_infrastructure-based_2020}}
\newcommand{\chen}{Chen and Schwertfeger 2019 \cite{chen_heterogeneous_2019}}
\newcommand{\esquivel}{Esquivel \etals 2007 \cite{esquivel_calibration_2007}}
\newcommand{\carrera}{Carrera \etals 2011 \cite{carrera_slam-based_2011}}
\newcommand{\pollok}{Pollok and Monari 2016 \cite{pollok_visual_2016}}
\newcommand{\lebraly}{L{\'e}braly \etals 2010 \cite{lebraly_calibration_2010}}
\newcommand{\robinson}{Robinson \etals 2017 \cite{robinson_robust_2017}}
\newcommand{\shahReview}{Shah \etals 2012 \cite{shah_overview_2012}}
\newcommand{\kalibrFull}{Furgale \etals 2013 \cite{kalibr}}
\newcommand{\maye}{May \etals 2013 \cite{maye_self-supervised_2013}}
\newcommand{\mayeThesis}{Maye 2014 \cite{maye_online_2014}}

Recent work on multi-camera and/or multi-sensor calibration can be divided into pattern- versus infrastructure-based methods. Other differences between methods include requirements about fully or partially shared fields of view, mobile or stationary camera systems, whether camera synchronization is needed, and how the rigid constraints of a multi-camera system are represented and computed.  Many works use existing solutions to the hand-eye and hand-eye, robot-world calibration methods, hand-eye, robot-world methods are reviewed in \shahReview.

{\bf Pattern methods.} Pattern-based methods require calibration patterns. \li \ describes a new planar pattern type that allows for partial pattern detection and pose estimation; non-overlapping views are possible, but at least two cameras need to see part of one pattern at each image acquisition. \liucaliber \ (Caliber) do not specify the pattern type; instead the user designates the calibration rig and scene via a kinematic tree. The final step of both \cite{li_bo_multiple-camera_2013} and \cite{liu_caliber:_2016} is reprojection error minimization.  

The Kalibr toolbox, \kalibrFull, \maye, \mayeThesis, estimates rigidity constraints between cameras via an information theoretic measure. Kalibr requires the use of a calibration pattern that can be seen by pairs of cameras; the implementation released by the authors uses April Tags \cite{olson2011tags}.

{\bf Infrastructure methods.} Infrastructure methods use the sensed environment to calibrate a multi-camera system, usually through an existing Structure-from-Motion framework, such as COLMAP \cite{schonberger_structure--motion_2016} or a version of SLAM. \esquivel \ and \lebraly \ create maps for individual cameras, then enforce rigidity constraints over time steps by closed-form hand-eye calibration solutions. \esquivel \ perform a global optimization for algebraic error, while \lebraly \ minimize for reprojection error via bundle adjustment that incorporates the camera rig's rigidity constraints. \carrera \ fuse maps from individual cameras and compute transformations between maps to estimate rigidity constraints, followed by a bundle adjustment with rigidity constraints similar to \cite{lebraly_calibration_2010}.

\lin \ is an infrastructure-based method that uses the 1D radial camera model \cite{thirthala_radial_2012} to estimate extrinsic parameters for each image in a sequence, up to an unknown forward translation (meaning the $z$ component of a translation vector is unknown). Camera rig transformations from the map to calibration rig are estimated in a local assign-and-estimate loop, followed by stages of non-linear least squares minimization to refine extrinsic parameters, determine camera intrinsics and forward translation. A final step minimizes for reprojection error and optimizes all parameters (intrinsics, extrinsics, and rig poses).

{\bf Hybrid methods.} Several methods use extra hardware for calibration in challenging settings. \mishra \ calibrate ceiling-mounted cameras using a mobile robot with its own camera and calibration pattern. \chen \ calibrate a multi-sensor system of cameras and 3D lidar units with the aid of patterns and a camera-based motion capture system. A hand-held camera estimates a map using SLAM to calibrate a surveillance camera network in \pollok. \robinson \ calibrate cameras with non-overlapping views by adding intermediate cameras such that a planar calibration pattern could be seen by pairs of cameras.

Our work, CALICO, formulates the calibration problem and rigidity constraints in a manner that is most similar to \esquivel \ and \lebraly \ in that the transformation between frames is estimated. Instead of a camera-centered gauge and infrastructure-based approach as in those works, we use an object-centered gauge and a pattern-based approach. CALICO also shares some similarities with \liucaliber \ in that multiple calibration targets may be used, and the technique is flexible in that stationary or mobile multi-camera systems may be calibrated.  Our work is different from \cite{liu_caliber:_2016} in that we do not require users specify of a kinematic tree; in CALICO, the relationships between patterns and cameras are estimated as part of the method.

\section{Multi-camera calibration problem formulation}
\label{overview}

The multiple-camera calibration problem is the computation of relative poses between cameras, or, camera poses with respect to a coordinate system. These poses are represented by homogeneous transformation matrices (HTMs): $4 \times 4$ matrices composed of a $3 \times 3$ orthogonal rotation matrix $\mathbf{R}$ and a column vector $\mathbf{t}$ of size $3$ representing the translation, or 
\begin{equation}
\begin{bmatrix} \mathbf{R} & \mathbf{t} \\
\mathbf{0}^T & 1 \end{bmatrix}.
\label{eq:htm}
\end{equation}

\begin{table}
	\caption{Notation used in this paper.}
	\begin{centering}
		\resizebox{0.48\textwidth}{!}
			{
			\begin{tabular}{@{}l|l@{}|l@{}}
				\hline
				 & Notation & Meaning\\
				\hline
				\hline
				1 & $n_c$ & Number of cameras \\
				\hline 
				2 & $n_p$ & Number of patterns \\
				\hline 
				3 & $n_t$ & Number of time labels \\
				\hline
				4 & $w$ & World coordinate frame \\
				\hline
				5 & $\mathcal{C} = \{c_0, c_1, \dots , n_c - 1 \}$ & Set of camera indices\\
				\hline
				6 & $\mathcal{P} = \{p_0, p_1, \dots , n_p - 1 \}$ & Set of pattern indices\\
				\hline
				7 & $\mathcal{T} = \{t_0, t_1, \dots , n_t - 1 \}$ & Set of time label indices\\
				\hline
				8 & $\mathbb{C} = \{{^{c_i}\mathbf{C}_w} \in SE(3) \vert {c_i \in \mathcal{C}} \}$ & \begin{tabular}{@{}l@{}} HTMs from world coordinate \\ frame to cameras\end{tabular}\\
				\hline
				9 & $\mathbb{T} = \{^{w @ t_j}\mathbf{T}_{w} \in SE(3)\vert {t_j \in \mathcal{T}} \}$ & \begin{tabular}{@{}l@{}} HTMs from world coordinate \\ frame to time labels\end{tabular}\\
				\hline
				10 & \begin{tabular}{@{}l@{}}  $\mathbb{P} = \{^{p_k @ t_j}\mathbf{P}_{w @ t_j} \in {SE(3)} \vert$\\
					  $\quad \; p_k \in \mathcal{P},\{t_j, t_m\} \in \mathcal{T}, $\\
				      $\quad \; {^{p_k @ t_j}\mathbf{P}_{w @ t_j}} = {^{p_k @ t_m}\mathbf{P}_{w @ t_m}} \}$ \end{tabular} & \begin{tabular}{@{}l@{}} HTMs from world coordinate \\ system to patterns\end{tabular}\\
				\hline
				11 & \begin{tabular}{@{}l@{}} $\mathbb{A} = \{{^{c_i @ t_j}\mathbf{A}^{w @ t_j}_{w @ p_k}} \in SE(3)\vert$\\
					$\quad \; c_i \in \mathcal{C}, p_k \in \mathcal{P}, t_j \in \mathcal{T} \}$ \end{tabular} & \begin{tabular}{@{}l@{}} HTMs relating pattern $p_j$\\to camera $c_i$ at time $t_k$.\end{tabular}\\
				\hline
				12 & ${^{c_i}\mathbf{C}} = {^{c_i}\mathbf{C}_w}$ & Abbreviated $\mathbb{C}$ notation\\
				\hline
				13 & ${^{t_j}\mathbf{T}} = {^{t_j}\mathbf{T}_w}$ & Abbreviated $\mathbb{T}$ notation\\
				\hline
				14 & ${^{p_k}\mathbf{P}} = {^{p_k @ t_j}\mathbf{P}_{w @ t_j}}$ & Abbreviated $\mathbb{P}$ notation\\
				\hline
				15 & ${^{c_i}\mathbf{A}^{t_j}_{p_k}} = {^{c_i @ t_j}\mathbf{A}^{w @ t_j}_{w @ p_k}}$ & Abbreviated $\mathbb{A}$ notation\\
				\hline
				16 & \multrow{$\mathbb{CS} = \{^{c_i}\mathbf{C} = {^{c_i}\mathbf{A}^{t_j}_{p_k}} \; {^{p_k}\mathbf{P}} \; {^{t_j}\mathbf{T}}$ \\
				$\quad \; c_i \in \mathcal{C}, p_k \in \mathcal{P}, t_j \in \mathcal{T} \}$ } & \multrow{Constraint set for \\ multi-camera calibration} \\
				\hline
				17 & $\CSp \subseteq \CS$ & \multrow{Constraint subset of $\CS$\\where all $\mathbb{C}$, $\mathbb{P}$, $\mathbb{T}$\\ variables are initialized}\\
				\hline
				18 & $r_{ae} \in [0, 1]$ & \multrow{proportion of constraints\\ added to the algebraic\\ error minimization.}\\
				\hline
				19 & $r_{rp} \in [0, 1]$ & \multrow{proportion of constraints\\ added to the reprojection\\ error minimization.}\\
				\hline
				20 & $ae$ & \multrow{Total algebraic error \\ over $\CS$}\\
				\hline
				21 & $re$ & \multrow{Total reprojection error\\ over $\CS$} \\
				\hline
				22 & $rrmse$ & \multrow{Root reprojection mean square\\ error over $\CS$}\\
				\hline 
				23 & $X$ & \multrow{3D point on a\\ calibration pattern}\\
				\hline 
				24 & $x$ & \multrow{2D image point representing\\ calibration pattern point $X$}\\
				\hline
				25 & $\mathbb{X}$ & \multrow{Set of world-to-image\\ calibration pattern\\ point pairs $(X,x)$}\\
				\hline 
				26 & $\mathbb{Y}$ & \multrow{Set of reconstructed \\ calibration pattern\\ points}  \\
				\hline 
			\end{tabular}
		}
	\end{centering}
	\label{table:symbols}
\end{table} 

We consider multi-camera calibration in a pattern-based, non-synchronized context. Camera poses are computed relative to a coordinate system; we assume internal camera calibration parameters are available or can be computed from an image dataset. Table \ref{table:symbols} shows the notation used in this paper. There are $n_c$ cameras, which are rigidly attached to each other, or rigid surfaces, and referred to as the \textbf{camera rig}. In our formulation, we want to compute the $n_c$ transformations from a world coordinate system to each camera, HTMs $\mathbb{C}$. 

There are $n_p$ calibration patterns rigidly attached to each other, or rigid surfaces; the pattern set is referred to as the \textbf{pattern rig}.\footnote{It may be more standard to refer to the pattern rig as the calibration rig; we chose a name that had a different first letter to reduce possible confusion with camera rig.} Each pattern in the set is distinguishable from other patterns, and images of a partial pattern can be used to estimate the camera pose with respect to the pattern. 

All cameras capture an image at each of $n_t$ time labels. Synchronized image acquisition is not required. After an image capture for all cameras, the pattern rig, camera rig, or both, are moved and images acquired to create data for the next time label. Variations to this data acquisition procedure are possible and discussed in Section \ref{sec:data_acqusition}.

\subsection{Representation of rigid constraints: constraint set $\CS$}
\label{ss:problem}

We define camera HTMs $\Cam$ with respect to a world coordinate system $w$ (discussed in Section \ref{ss:w}). We use extra unknown variables, pattern and time HTM sets $\Pat$ and $\Tim$. Variables from $\Cam$, $\Pat$, and $\Tim$ are indexed by $\mathcal{C}$, $\mathcal{P}$, and $\mathcal{T}$, respectively. $\Pat$'s HTMs represent the transformations between the world coordinate system $w$ and the pattern rig's individual patterns; $\mathbb{T}$'s HTMs represent the transformations between the  world coordinate system $w$ to its pose at different time labels. While in the formal listing of $\mathbb{P}$ in Table \ref{table:symbols}, line 10, time sub- and superscripts are included, $\vert \mathbb{P} \vert = \vert \mathcal{P} \vert$; there is one HTM in $\mathbb{P}$ for every pattern in the pattern rig.

We also use observation HTMs $\mathbb{A}$ in our formulation; $\mathbb{A}$'s HTMs represent the pose of a camera with respect to a pattern at a particular time label. In other words, when camera $c_i$ views pattern $p_k$ at time $t_j$, the transformation from the coordinate system defined by pattern $p_k$ to camera $c_i$ is estimated through Perspective-n-Point (PnP) pose computation as is typical in single-camera calibration. The resulting pose is notated as ${^{c_i @ t_j}\mathbf{A}^{w @ t_j}_{w @ p_k}}$.

With the unknown variables $\mathbb{C}$, $\mathbb{P}$, and $\mathbb{T}$, and observations $\mathbb{A}$, we create a set of constraints, $\CS$, with form
\begin{equation}
{^{c_i}\mathbf{C}_w} = {^{c_i @ t_j}\mathbf{A}^{w @ t_j}_{w @ p_k}}\;{^{p_k @ t_j}\mathbf{P}_{w @ t_j}} \; {^{t_j}\mathbf{T}_w}.
\label{eq:generic-constraint}
\end{equation}
and this constraint is represented in a graph in Figure \ref{fig:foundational_relationship}.

The notation above is quite long, so we use abbreviated forms for the remainder of this paper (Table \ref{table:symbols}, lines 12-15). Then Eq. \ref{eq:generic-constraint} is written as
\begin{equation}
{^{c_i}\mathbf{C}} = {^{c_i}\mathbf{A}^{t_j}_{p_k}} \; {^{p_k}\mathbf{P}} \; {^{t_j}\mathbf{T}}.
\label{eq:abbrev-generic-constraint}
\end{equation}

\begin{figure}[t]
	\centering
	\includegraphics[width=1\linewidth]{includes/scenario_general_single.tikz}  
	\caption{Graphical representation of one constraint in $\CS$.}
	\label{fig:foundational_relationship}
\end{figure}

{\bf Example 1.} Consider a subset of constraints from a multi-camera calibration problem in Eq.s \ref{eq:constraintsa}-\ref{eq:constraintsd}. Notice that camera $c_0$ has one observation at time $t_0$ in Eq. \ref{eq:constraintsa} and two observations at $t_1$ in Eqs. \ref{eq:constraintsb} and \ref{eq:constraintsc}, as two patterns ($p_0$, $p_1$) were within camera $c_0$'s field of view and consequently two observations are generated. On the other hand, camera $c_1$ only observed $p_1$ at time $t_1$.
\begin{align}
{}^{c_0}\mathbf{C} = {}^{c_0}\mathbf{A}_{p_0}^{t_0}\; {}^{p_0}\mathbf{P} \; {^{t_0}}\mathbf{T} \label{eq:constraintsa}\\
{}^{c_0}\mathbf{C} = {}^{c_0}\mathbf{A}_{p_0}^{t_1}\; {}^{p_0}\mathbf{P} \; {^{t_1}}\mathbf{T} \label{eq:constraintsb}\\
{}^{c_0}\mathbf{C} = {}^{c_0}\mathbf{A}_{p_1}^{t_1}\; {}^{p_1}\mathbf{P} \; {^{t_1}}\mathbf{T} \label{eq:constraintsc}\\
{}^{c_1}\mathbf{C} = {}^{c_1}\mathbf{A}_{p_1}^{t_1}\; {}{^{p_1}\mathbf{P}} \; {^{t_1}}\mathbf{T} \label{eq:constraintsd}
\end{align}

\begin{figure*}
	\centering
	\includegraphics[width=0.7\linewidth]{includes/multi1-edit-colors.tikz}  
	\caption{The interaction graph for Dataset Mult-1. Edges connected to Camera ${}^{0}\mathbf{C}$ are blue, those connected to Camera ${}^{1}\mathbf{C}$ are light grey.  $\mathbf{T}^{0} = t^*$ and ${}^{0}\mathbf{P} = p^*$ in this dataset.}
	\label{fig:interaction-graphs}
\end{figure*}

\subsection{Non-linear cost functions: algebraic and reprojection error}\label{ss:cost-functions}
With Eq. \ref{eq:abbrev-generic-constraint} generated for every pattern detection in the data set, we consider the multi-camera calibration problem as an optimization problem: with initial solutions, we minimize the algebraic error of the constraint set (Eq. \ref{eq:generic-algebraic-error-cost}), and then refine the estimates for $\mathbb{C}$, $\mathbb{P}$, $\mathbb{T}$ by minimizing for reprojection error.  

{\it Algebraic error.} The algebraic error cost function represents the difference between the left and right sides of a constraint, summed over all constraints in $\CS$, 
\begin{equation}
\argmin\limits_{\mathbb{C}, \mathbb{P}, \mathbb{T}}\sum_{ \mathbb{CS}} \left\Vert {^{c_i}\mathbf{C}} - {^{c_i}\mathbf{A}^{t_j}_{p_k}} \; {^{p_k}\mathbf{P}} \; {^{t_j}\mathbf{T}} \right\Vert^2_{{F}},
\label{eq:generic-algebraic-error-cost}
\end{equation}
where $\left\Vert \cdot \right\Vert_{F}$ is the Frobenius norm.

{\it Reprojection error.} ${^{c_i}\mathbf{A}^{t_j}_{p_k}}$, estimated from image data, does not account for the problem's rigidity constraints. For the reprojection error cost function, we generate a HTM ${^{c_i}\hat{\mathbf{A}}^{t_j}_{p_k}}$ that represents the HTM relating pattern $p_k$ to camera $c_i$ at time $t_j$: 
\begin{equation}
{^{c_i}\hat{\mathbf{A}}^{t_j}_{p_k}} = ^{c_i}\mathbf{C} \; {^{t_j}\mathbf{T}}^{-1} \; {^{p_k}\mathbf{P}} ^{-1},
\label{eq:rigid-constraint-proj-mat}
\end{equation}
similarly as \cite{tabb_parameterizations_2015,tabb_solving_2017}. 

A three-dimensional point $X$ on a calibration pattern and the corresponding two-dimensional point $x$ in the image have the projective relationship in homogenous coordinates,
\begin{equation}
x = {}^{c_i}\hat{\mathbf{A}}^{t_j}_{p_k} X.
\label{eq:image-world-relationship}
\end{equation}

The detected image point that corresponds to $X$ is $\tilde{x}$, and its reprojection error is $\left\Vert{x - \tilde{x}}\right\Vert^2$. We substitute and combine Eq.s \ref{eq:rigid-constraint-proj-mat} and \ref{eq:image-world-relationship} to represent total reprojection error over $\mathbb{CS}$:
\begin{equation}
re =\sum_{\mathbb{CS}}\sum_{ \left(X,\tilde{x} \right) \in \mathbb{X}} \left\Vert{}^{c_i}\mathbf{C}  \left({^{t_j}\mathbf{T}}\right)^{-1} \left({{}^{p_k}\mathbf{P}}\right)^{-1} X - {\tilde{x}}\right\Vert^{2},
\label{eq:reprojection error}
\end{equation}
where $\mathbb{X}$ is the set of world-to-image calibration pattern point pairs $(X,x)$ detected in one constraint in $\mathbb{CS}$.

As with the algebraic error, we represent the reprojection error as a minimization problem as well,
\begin{equation}
\argmin\limits_{\mathbb{C}, \mathbb{P}, \mathbb{T}}\sum_{\mathbb{CS}}\sum_{ \left(X,\tilde{x} \right) \in \mathbb{X}} \left\Vert{}^{c_i}\mathbf{C}  \left({^{t_j}\mathbf{T}}\right)^{-1} \left({{}^{p_k}\mathbf{P}}\right)^{-1} X - {\tilde{x}}\right\Vert^{2}.
\label{eq:reprojection error-opti}
\end{equation}

\subsection{Gauge freedom of world coordinate frame $w$}\label{ss:w}
The constraint set $\CS$ is indeterminate with respect to $w$. We fix the world coordinate system $w$ to the coordinate system of a pattern $p^* \in \mathcal{P}$ at time label $t^* \in \mathcal{T}$, such that $^{p^*}\mathbf{P} = \mathbf{I}_4$ and  $^{t^*}\mathbf{T} = \mathbf{I}_4$, where $\mathbf{I}_4$ is an identity matrix of size four. Our heuristic for choosing $p^*$ and $t^*$ is described in Section \ref{ss:choose}. This is an example of an {\it object-centered} gauge \cite{triggs_bundle_2000}.

{\bf Example 2.} Specifying $w$ simplifies some of the constraints. Suppose $p^* = p_0$ and $t^* = t_1$. The example from above becomes 
\begin{alignat}{2}
{}^{c_0}\mathbf{C} &= {}^{c_0}\mathbf{A}_{p_0}^{t_0}\; \mathbf{I}_4 \; {^{t_0}}\mathbf{T} &&\rightarrow {}^{c_0}\mathbf{C} = {}^{c_0}\mathbf{A}_{p_0}^{t_0}  {^{t_0}}\mathbf{T} \label{eq:constraintse}\\
{}^{c_0}\mathbf{C} &= {}^{c_0}\mathbf{A}_{p_0}^{t_1}\; \mathbf{I}_4 \; \mathbf{I}_4 &&\rightarrow {}^{c_0}\mathbf{C} = {}^{c_0}\mathbf{A}_{p_0}^{t_1}\label{eq:constraintsf}\\
{}^{c_0}\mathbf{C} &= {}^{c_0}\mathbf{A}_{p_1}^{t_1}\; {}^{p_1}\mathbf{P} \; \mathbf{I}_4 &&\rightarrow {}^{c_0}\mathbf{C} = {}^{c_0}\mathbf{A}_{p_1}^{t_1}\; {}^{p_1}\mathbf{P} \label{eq:constraintsg}\\
{}^{c_1}\mathbf{C} &= {}^{c_1}\mathbf{A}_{p_1}^{t_1}\; {}{^{p_1}\mathbf{P}} \; \mathbf{I}_4 &&\rightarrow {}^{c_1}\mathbf{C} = {}^{c_1}\mathbf{A}_{p_1}^{t_1}\; {}{^{p_1}\mathbf{P}}. \label{eq:constraintsh}
\end{alignat}
Since ${}^{c_0}\mathbf{A}_{p_0}^{t_1}$ is an observation and known, ${}^{c_0}\mathbf{C}$ can be initialized, and so on through the remaining constraints.

\section{Method description} 
\label{sec:estimation}

\begin{algorithm}
	\caption{CALICO: Multi-camera calibration method.} \label{alg:calibration}
	\begin{algorithmic}[1]
		\Require{Set of constraints, $\CS$.}
		\Ensure{Set of HTMs $\Cam$, $\Pat$, $\Tim$.} 
		\State {Determine intrinsic and extrinsic camera parameters with respect to visible patterns at all time labels, \ref{ss:intrinsic-cali}.}
		\State {Test if the multi-camera system can be calibrated, \ref{sec:graph_test}.}
		\State {Choose the gauge; find $p^*$ and time $t^*$, \ref{ss:choose}.} 
		\While {Any variable can be initialized}
		\State {Initialize individual, then variable pairs, \ref{ss:initial-solution}.}
		\State {Minimize algebraic error for all initialized variables, \\ \ \ \ \ \ Eq. \ref{eq:generic-algebraic-error-cost}, every $r_{ae}$ iterations. }
		\EndWhile
		\State {Refine HTMs $\Cam$, $\Pat$, $\Tim$ by minimizing reprojection error, Eq. \ref{eq:reprojection error-opti}; add batches of $r_{rp}$ variables to the minimization problem.}
	\end{algorithmic}
\end{algorithm}

We want to compute $\mathbb{C}$, with rigidity constraints represented by $\mathbb{P}$, $\mathbb{T}$ which are also unknown. We ultimately want to minimize reprojection error, Eq. \ref{eq:rigid-constraint-proj-mat}. A brief sketch of the method is: fix the gauge to eliminate indeterminacy. Then, we initialize the values of unknown variables similarly to Example 2, using a local search; we find those constraints that have only one uninitialized variable remaining, and solve with a closed-form method. When no more constraints with only one variable initialized remain, we examine constraints with two uninitialized variables, and then solve for two variables with a closed-form method. We find approximate solutions to the non-linear least squares problems for algebraic error, Eq. \ref{eq:generic-algebraic-error-cost} during the search-and-solve procedure. Finally, we minimize for reprojection error, Eq. \ref{eq:reprojection error-opti}, over the entire set of variables $\CS$. 

Before these steps, though, we compute the internal and external camera calibration (${^{c_i}\mathbf{A}^{t_j}_{p_k}}$) for all cameras in the dataset. We test whether CALICO can calibrate the dataset. Alg. \ref{alg:calibration} describes the steps of the method. 

\subsection{Individual camera calibration}
\label{ss:intrinsic-cali}

Individual cameras are calibrated for intrinsic and extrinsic camera calibration parameters. Each pattern detection triggers the generation of one constraint equation in $\mathbb{CS}$, Eq. \ref{eq:abbrev-generic-constraint}. The extrinsic parameters -- the HTMs from a pattern's world coordinate system to the camera coordinate system -- make up $\mathbb{A}$. Some images may contain pattern detections of more than one pattern at one time label; in these cases a constraint equation is generated and HTM ${^{c_i}\mathbf{A}^{t_j}_{p_k}}$ for each pattern detected. We use OpenCV's implementation of Zhang's calibration algorithm \cite{zhang_flexible_2000, bradski_opencv_2000} for this step.  

\subsection{Test ability to calibrate}
\label{sec:graph_test}

We construct an {\it interaction graph} whose vertices are the variables in $\lbrace \mathbb{C} \cup \mathbb{T} \cup \mathbb{P} \rbrace$. For each constraint in $\mathbb{CS}$, edges are created between the variables in that constraint. We then compute the number of connected components in the graph. The entire camera rig can be calibrated with CALICO by a particular image dataset if there is exactly one connected component. An example from dataset Mult-1 is shown in Figure \ref{fig:interaction-graphs}.

If the graph consists of multiple connected components, then the cameras corresponding to each component can be calibrated with respect to each other in the connected component but not with respect to cameras in different connected components. 

\subsection{Choose gauge}
\label{ss:choose}

We choose the gauge and define the world coordinate system $w$ by a pattern $p^* \in \mathcal{P}$ at time label $t^* \in \mathcal{T}$, such that the greatest number of variables can be initialized.  The time and pattern combination with the greatest frequency in $\mathbb{CS}$ is chosen as $t^*$ and $p^*$, i.e.,
\begin{multline} 
p^{*}=\\
\argmax\limits_{p_k}\!\!\!\sum_{p_k \in \mathcal{P}} \Big|\left \{^{c_i}\mathbf{C}\!=\!{^{c_i}\mathbf{A}^{t_j}_{p_k}}{^{p_k}\mathbf{P}}{^{t_j}\mathbf{T}}
\, | c_i \in \mathcal{C},  t_j \in \mathcal{T} \right \} \in \mathbb{CS} \Big|,
\label{eq:exemplar_target}
\end{multline}
or the pattern that has been observed the most times by any camera at any time label, and
\begin{multline} 
t^{*}=\\
\argmax\limits_{t_j}\sum_{t_j \in \mathcal{T}} \Big|\left \{^{c_i}\mathbf{C} = {^{c_i}\mathbf{A}^{t_j}_{p^*}}{^{p^*}\mathbf{P}}{^{t_j}\mathbf{T}}
\; | c_i \in \mathcal{C} \right \} \in \mathbb{CS} \Big|,
\label{eq:exemplar_time}
\end{multline}
which is the time label corresponding to the largest number of observations of pattern $p^{*}$ by any camera. The two HTMs corresponding to indices $p^*$ and $t^*$ are initialized to ${}^{p^*}\mathbf{P}= \mathbf{I}_4$ and ${}^{t^*}\mathbf{T} = \mathbf{I}_4$.  

\subsection{Initial solutions search and non-linear least squares minimization refinement}
\label{ss:initial-solution}

Initial solutions for $\Cam$, $\Pat$, and $\Tim$ are found through a local search-and-solve procedure (Sections \ref{sss:closed-form-one} and \ref{sss:closed-form-two}). We first solve constraints involving only one unknown variable and then two. Then, subsets of $\CS$ are used to create cost functions that approximately minimize non-linear least squares costs functions of algebraic and reprojection error, which further refines variables' estimates. 

First, we search through $\mathbb{CS}$ for constraints where $p_k=p^*$ and $t_j=t^*$ and initialize the camera HTMs for those constraints, as in Example 2. The number of initialized variables is at least three because of the way in which $p^*$ and $t^*$ were chosen; there is at least one constraint where $p_k=p^*$ and $t_j=t^*$ and consequently one camera variable that may be initialized.  

Following the first search, we iterate through $\mathbb{CS}$ and handle constraints with only one uninitialized variable remaining. The variable's HTM is initialized using closed-form methods discussed in Section \ref{sss:closed-form-one}. When there are no more constraints in $\CS$ having only one uninitialized variable, constraints having two uninitialized variables are solved using methods for robot-world, hand-eye calibration, discussed in Section \ref{sss:closed-form-two}. Then we return to iterating through $\mathbb{CS}$ and handling constraints with only one uninitialized variable. We refer to the set of constraints where all three variables are initialized as $\CSp$.

Already-initialized variables' values are refined by minimization of algebraic error at several points during the search for uninitialized variables. Finally, we minimize reprojection error.

\subsection{Initializing a single variable} \label{sss:closed-form-one}

Let $\CS^{(m)}$ be the set of elements of $\CS$ for which only one HTM variable $\mathbf{X^{(m)}}$ is unknown (at a given iteration, $\mathbf{X}^{(m)}$ could be either ${}^{c_i}\mathbf{C}$, ${}^{p_k}\mathbf{P}$, or $^{t_j}\mathbf{T}$). We solve Eq. \ref{eq:abbrev-generic-constraint} for $\mathbf{X}^{(m)}$ by rearranging the terms to the form
\begin{equation}
\mathbf{X}^{(m)}\mathcal{A} = \mathcal{B}, 
\label{eq:inv_sol}
\end{equation}
where $\mathcal{A}$ and $\mathcal{B}$ are HTMs that are data (from $\mathbb{A}$), initialized variables, or multiplied combinations of HTMs from $\mathbb{A}$ and initialized variables within a constraint. $\mathbf{X}^{(m)}$ is the unknown transformation. If $|\CS^{(m)}| = 1$, we simply compute $\mathbf{X}^{(m)} = (\mathcal{A})^{-1}\mathcal{B}$. Otherwise, we solve for $\mathbf{X}^{(m)}$ with $\CS^{(m)}$ using Shah's closed-form method for registering two sets of six degrees-of-freedom data \cite{shah_comparing_2011}.

$\mathbf{X}^{(m)}$'s initial solution is refined by finding an approximate solution to 
\begin{equation}
\argmin\limits_{\mathbf{X}^{(m)}}\sum_{\CS^{(m)}} \left\Vert\mathbf{X}^{(m)}\mathcal{A} - \mathcal{B}\right\Vert^2_{{F}}
\label{eq:refine-single}
\end{equation}
using the Levenberg-Marquardt algorithm.

\subsection{Initializing variable pairs}\label{sss:closed-form-two}

Similarly, let $\CS^{(m)}$ be the set of elements of $\CS$ HTMs for which both $\mathbf{X}^{(m)}$ and $\mathbf{Z}^{(m)}$ are unknown. In this case, a constraint with two unknown variables can be rearranged into the form of the robot-world, hand-eye calibration problem 
\begin{equation}
\mathcal{A}\mathbf{X^{(m}} = \mathbf{Z^{(m)}}\mathcal{B},
\label{eq:rwhec}
\end{equation}
where $\mathcal{A}$ and $\mathcal{B}$ are known HTMs.  We solve for the unknown variables using Shah's closed-form method for the robot-world, hand-eye problem \cite{shah_solving_2013}. In some cases, $\mathcal{A}$ or $\mathcal{B}$ may be identity matrices. As with single variables, we refine this initial estimate by finding an approximate solution to
\begin{equation}
\argmin\limits_{\mathbf{X}^{(m)},\mathbf{Z}^{(m)}}\sum_{\CS^{(m)}} \left\Vert\mathcal{A}\mathbf{X}^{(m)} - \mathbf{Z}^{(m)}\mathcal{B}\right\Vert^2_{{F}}
\label{eq:refine-pair}
\end{equation}
using the Levenberg-Marquardt algorithm.

\subsection{Variable initialization order}

At each iteration, there may be many choices of single variables or pairs of variables to solve for via the methods of sections \ref{sss:closed-form-one} and \ref{sss:closed-form-two}. Solving for single variables is always selected over solving for variable pairs. Beyond this choice, the variable initialization order is determined by a heuristic that prioritizes variables contained in the largest number of remaining constraints. That is, we select the HTM $\mathbf{X}^{(m)}$ that maximizes $\left| \CS^{(m)} \right|$. Ties are broken by choosing $\Cam$ variables first, then $\Pat$, and finally $\Tim$, and breaking further ties by choosing the variable with the smallest index first.

\subsection{Minimization of algebraic error over $\CSp$ }\label{sss:ae-mini}

The variables' estimates are refined by minimizing for algebraic error, Eq. \ref{eq:generic-algebraic-error-cost}, over the set $\CSp$, using the Levenberg-Marquardt algorithm. This minimization is done after a user-specified proportion of iterations, $r_{ae}$, through the {\bf while} loop at line 5 of Algorithm \ref{alg:calibration}. For example, if $r_{ae}$ is $0.20$ and there are $100$ constraints in $\CS$, minimization of Eq. 3 over $\CSp$ occurs when the number of passes through initializing one variable (\ref{sss:closed-form-one}) or variable pairs (\ref{sss:closed-form-two}) is 20, 40, 60, and so on.

\subsection{Final refinement: minimization of reprojection error.}

Finally, the variables' estimates are refined by minimizing for reprojection error, Eq. \ref{eq:reprojection error-opti} using the Levenberg-Marquardt algorithm.. Similarly to the minimization of algebraic error, the reprojection error is refined over subsets of $\CSp$. We use a user-specified proportion of constraints, $r_{rp}$, to control when and how the variables are minimized. If $r_{rp} = 0.5$, then reprojection error is minimized for all variables in half of $\CS$; then the constraints in the second half of $\CS$ are added to the cost function and the minimization performed.


\section{Experiments}

CALICO was evaluated in simulated and real-world experiments. First, we describe experimental details in Section \ref{sec:data_acqusition}, introduce three evaluation metrics in Section \ref{ss:eval}, and then describe datasets and results in Sections \ref{ss:datasets} and \ref{sec:results}, respectively.  We use the implementation of the Levenberg-Marquardt method found in the software package Ceres \cite{ceres-solver}. The results shown in this paper were generated on a workstation with one 12-core 2.70GHz processor. OpenCV \cite{bradski_opencv_2000} was used to compute the camera calibration parameters.

The parameter, $r_{ae}$, indicating the proportion of variables included in subsequent runs of algebraic error minimization (Eq. \ref{eq:generic-algebraic-error-cost}) was set to 0.2 for all datasets.  The parameter indicating the proportion of variables included in subsequent runs of the reprojection error minimization, $r_{rp}$ was set to 0.5 for all datasets (Eq. \ref{eq:reprojection error}).

\subsection{Evaluation}
\label{ss:eval}

We used three metrics for evaluation: algebraic error, reprojection root mean squared error, and reconstruction accuracy error.  

\subsubsection{Algebraic error}

The algebraic error represents the fit of the estimated HTMs to $\CS$. It is given by
\begin{equation}
ae=\frac{1}{|\CS|}\sum_{\CS} \left\Vert {}^{c_i}\mathbf{C} - {}^{c_i}\mathbf{A}_{p_k}^{t_j} {{}^{p_k}\mathbf{P}} {}^{t_j}{\mathbf{T}} \right\Vert^2_{{F}}. 
\label{eq:algebraic_error}
\end{equation}

\subsubsection{The Reprojection Root Mean Squared Error (rrmse)} 
The reprojection root mean squared error is
\begin{equation}
rrmse= \sqrt{\frac{1}{N}re} \, ,
\label{eq:rrmse}
\end{equation}
where $N  = \sum_{ \CS }{\left| \mathbb{X} \right|}$ is the total number of calibration pattern points observed and $re$ is reprojection error from Equation \ref{eq:reprojection error}.

\subsubsection{Reconstruction Accuracy Error}

Reconstruction accuracy error, $rae$, is used here in a similar way as in \cite{tabb_solving_2017}, to assess the method's ability to reconstruct the three-dimensional location of calibration pattern points. Given detections of a pattern point in images over multiple cameras and times, the three-dimensional point that generated those pattern points is estimated by N-view triangulation.  Reconstruction accuracy error ($rae$) is the Euclidean difference between estimated and ground truth world points. The ground truth world points consist of the coordinate system defined by the calibration pattern, so these point are known even in real settings.  

The three-dimensional point $X$ that generated the corresponding image points $x$ can be found by solving the following minimization problem
\begin{equation}
\hat{\mathcal{Y}} = \argmin\limits_{X}\sum_{ \CS} \left\Vert{}^{c}\mathbf{C}  \left({\mathbf{T}^{t}}\right)^{-1} \left({{}^p\mathbf{P}}\right)^{-1} X - {x}\right\Vert^{2}.
\label{eq:min_world_point}
\end{equation}
Since Eq. \ref{eq:min_world_point} is a non-linear least squares problem, we require an initial solution for $X$. We use the direct linear transform (DLT) algorithm described in \cite{hartley_triangulation_1997} as the `Linear-LS Method' and the normalization method presented in \cite{hartley_defense_1997}. This approach to triangulation is also described in \cite{hartley_multiple_2004}. 

Using an initial solution computed using the DLT method, we find an approximate solution for $\hat{\mathcal{Y}}$ in Eq. \ref{eq:min_world_point} using the Levenberg-Marquardt algorithm.
$\hat{\mathcal{Y}}$ is found for all calibration pattern points detected in two or more constraints, generating the set $\mathbb{Y}$. The reconstruction accuracy error ($rae$) is the Euclidean distance between the estimated $\hat{\mathcal{Y}}_q$ points and corresponding calibration object points $X_q$. We report the mean $rae$ value over the set $\mathbb{Y}$:
\begin{equation}
rae =  \frac{1}{|\mathbb{Y}|}\sum_{\mathcal{Y}_{q} \in \mathbb{Y}}{\left\Vert \hat{\mathcal{Y}}_{q} - {X}_{q} \right\Vert}.
\label{eq:rae}
\end{equation}

\subsection{Comparison with Kalibr}
\label{ss:kalibr-comparison}

Part of our evaluation of CALICO includes a comparison with multi-sensor calibration method Kalibr.  We use aruco markers in our laboratory, but Kalibr uses April tags \cite{olson2011tags}, so we created simulated datasets with April tags for comparing Kalibr with CALICO. Another difference is that Kalibr requires one and only one calibration pattern, which sometimes imposes constraints on camera placement. 

Our philosophy on comparison with Kalibr is that we are not interested in very small differences in accuracy between our method and Kalibr. There are some situations where Kalibr is not able to calibrate a dataset, and CALICO is. Since we are using others' implementation, it is difficult to determine which characteristics of Kalibr calibrations are features of Kalibr as a theoretical method, and Kalibr's implementation, and extensively engineering another's implementation is beyond the scope of this paper.

Consequently, our comparison of CALICO with Kalibr is focussed on whether CALICO's calibrations are equivalent to the comparable state-of-the-art method, Kalibr, in the situations where Kalibr is able to calibrate, and not on whether CALICO `beats' Kalibr, and vice versa.

\subsection{Datasets}
\label{ss:datasets}

Our experiments include simulated and real datasets. They can be divided into a four types: \boxIt, \robot, \stereo, and \wbs. The \boxIt \ type has cameras arranged on the perimeter of the box and pointing towards the interior of the box. The \robot \ type consists of cameras lying on a cylinder, circle, or line and pointing outward. Stereo and wide-baseline stereo datasets have a pair of cameras; in wide-baseline stereo, individual cameras may view exclusive sets of individual aruco or April Tag pattern indices. Examples of these different dataset types are in Figure \ref{fig:simulatedDatasets}.

The datasets use either aruco markers \cite{garrido-jurado_automatic_2014} or April tags \cite{olson2011tags}. The real and simulated datasets are described in Tables \ref{table:description} and \ref{table:sim-description}, respectively.

\subsection{Pattern and camera configurations, data acquisition, and implementation details}
\label{sec:data_acqusition}

We used a set of planar calibration targets created with chessboard-type aruco tags \cite{garrido-jurado_automatic_2014} generated using OpenCV \cite{bradski_opencv_2000}, referred to as charuco patterns. The particular arrangement, and orientation, of the patterns is computed by CALICO. If a particular calibration pattern's orientation can be determined and its pattern index detected, there is no restriction on the type of pattern used if the connections between individual calibration patterns is rigid. For example, one could mount patterns on laboratory walls to satisfy rigidity requirements. One could also use a single pattern as is typical for camera calibration if the camera's field of view is shared.

Our real experiments' data acquisition process is as follows.  First, multiple images are acquired per camera to allow for internal parameter calibration.  Then, the user places the calibration rig in view of at least one camera.  The user indicates that the current acquisition is time label $t_0$ and acquires an image from all cameras.  Then, the calibration rig is moved such that at least one or more cameras view a pattern, not necessarily the same pattern, the user indicates that the time label and images are written from all of the cameras. This process is continued until $n_t$.

The use of `time label $t_0$' does not imply that the cameras are synchronized, but instead that the images are captured and labeled with the same time tag for the same position of the camera rig and pattern rigs.  A mechanism for doing so may be implemented through a user interface that allows the user to indicate that all cameras should acquire images, assign them a common time tag, and report to the user when the capture is concluded. 

Camera calibration for internal parameters is performed for each of the cameras independently. Individual patterns are uniquely identified through aruco or charuco tags \cite{garrido-jurado_automatic_2014}; cameras' extrinsic parameters (rotation and translation) are computed with respect to the coordinate systems defined by the patterns recognized in the images.  If two (or more) partial patterns are detected in the same image, that camera will have two (or more) extrinsic transformations defined for that time label, one for each of the patterns detected. 

The minimal requirement on the number and type of images to collect is that the interaction graph consist of a single connected component (Section \ref{sec:graph_test}). In real calibration scenarios, we found that larger values of $n_t$ produce better quality initial solutions than smaller values of $n_t$, as the quality of the initial solution (Section \ref{ss:initial-solution}) benefits from additional constraints between variables. The solution quality produced with CALICO can be assessed via the error metrics we describe in the next section; if the error is too high, one can acquire more images, add them to the dataset, and recalibrate.

\begin{table} 
	\caption{Real dataset descriptions. These datasets use charuco patterns.}
	\begin{center}
		{
			{
				\begin{tabular}{cccccc}
					\hline
					Dataset & type & $|\CS|$ & cameras & patterns & times ($n_t$) \\
					\hline
					Net-1 & \boxIt & 107 & 12 & 2 & 10 \\
					\hline
					Net-2 & \boxIt & 211 & 12 & 2 & 20 \\
					\hline
					Mult-1 & \robot & 20 & 2 & 2 & 10 \\
					\hline
					Mult-2 & \robot & 20 & 2 & 2 & 10 \\
					\hline
					Mult-3 & \robot & 72 & 4 & 3 & 24 \\
					\hline
				\end{tabular}
			}
		}
	\end{center}
	\label{table:description}
\end{table}

\subsubsection{Simulated experiments}
\label{ss:simulated}

\begin{table}
	\caption{Simulated dataset descriptions. Datasets Sim-1 and Sim-2 use charuco patterns, while the remainder use April tags.}
	\resizebox{0.48\textwidth}{!}
	{
	\begin{tabular}{llrrrrr}
		\hline
		Dataset & type & \multrow{max.\\distance\\between\\cameras} & $|\CS|$ & cameras & patterns & \multrow{times\\ ($n_t$)} \\
		\hline
		Sim-1 & \boxIt & 3162.3 & 87  & 8  & 4  & 43 \\
		\hline
		Sim-2 & \boxIt & 1677.1 & 472  & 16  & 4  & 37 \\
		\hline
		Sim-3-april & \stereo & 103.08  & 81 & 2 & 1 & 41 \\
		\hline
		Sim-4-april & \robot$^a$ & 210.81  & 665 & 12 & 1 & 100 \\
		\hline
		Sim-5-april & \robot$^a$ & 210.81 & 659 & 12 & 1 & 100 \\
		\hline
		Sim-6-april & \stereo & 375 & 109 & 2 & 1 & 55 \\
		\hline
		Sim-7-april & \robot$^b$ & 750 & 45 & 6 & 1 & 20 \\
		\hline
		Sim-8-april & \multrow{wide\\baseline\\stereo$^c$} & 1120.8 & 23 & 2 & 1 & 12 \\
		\hline
		Sim-9-april & \multrow{wide\\baseline\\stereo$^c$} & 1120.8 & 23 & 2 & 1 & 12 \\
		\hline
	\end{tabular}
}
	\bigskip
	
	\resizebox{0.48\textwidth}{!}
	{
		\begin{tabular}{ll}
			$^a$ & The camera poses are the same for these two datasets; difference is in the pattern poses.\\
			$^b$ & The six cameras lie on a line, with two cameras pointing towards $-z$,\\
			&  and the remaining four cameras pointing at $\pm 45$ and $\pm 90$ degrees from $-z$.\\
			$^c$ & The camera poses are the same for these two datasets; in Sim-8-April, there is a partition \\
			&  in between the two cameras such that each camera views half of the pattern's April Tags.\\
		\end{tabular}
	}
	\label{table:sim-description}
\end{table}

Simulated experiments were generated to allow assessment of CALICO to a ground truth, as well as to allow comparison to Kalibr. The simulated experiments include \boxIt, \robot, \stereo, and \wbs \ camera arrangements and are described in Table \ref{table:sim-description}. We used OpenGL to generate datasets. 

Datasets Sim-1 and Sim-2 use charuco patterns are of the \boxIt \ type, representing arrangements similar to those used in motion-capture experiments, where cameras are mounted on the walls around a room.  These datasets differ in camera density, number of cameras, distance from the pattern, and the resulting number of constraints.

The Kalibr multi-camera calibration method requires one and only one calibration pattern and its implementation provided by the authors require the use of April tags. Datasets Sim-3-april through Sim-9-april use April tags and use only one pattern. Sim-3-april is a two-camera simulated dataset; the cameras are roughly side-by-side, so it is of the \stereo \ type. In Sim-4-april, there are twelve cameras in three rows of four each, arranged around one quarter of a cylinder and pointing away from the cylinder's center axis. The calibration pattern is moved at 100 time steps in front of the cameras. Sim-5-april has the same arrangement of cameras as Sim-4-april, but random perturbations have been added to the pattern poses. 

Sim-7-april is a \robot \ type dataset, with camera arranged on a line. A criticism of CALICO has been that camera could be calibrated in some experiments by using a large enough pattern and Kalibr. Sim-8-april and Sim-9-april were generated to explore this criticism.  Sim-8-april is a \wbs \ type dataset, with a very large pattern such that most images contain a portion of the pattern.  Sim-9-april is the same dataset as Sim-8-april, except that there is a barrier such that one camera does not contain the set of April tags that are in the other cameras' images, and vice versa. While we wish to address the prior criticism of CALICO and explore Kalibr's performance in these situations, we reiterate our comparison philosophy from \ref{ss:kalibr-comparison} that it may be possible to alter Kalibr's implementation such that Kalibr can calibrate these situations, but that kind of work is beyond the scope of this paper.

\subsubsection{Stationary multi-camera system}
\label{ss:camera_network}

A multi-camera system was constructed using low cost webcameras, and arranged on two sides of a metal rectangular prism.  The pattern rig is constructed of two charuco patterns rigidly attached to each other.  To create datasets Net-1 and Net-2, the pattern rig was moved within the workspace. The stationary multi-camera system creates \boxIt \ datasets.

\newcommand{\clipleft}{2.5} 
\newcommand{\cliptop}{2}
\newcommand{\clipright}{2.5}
\newcommand{\clipbottom}{2}

\newcommand{\relwidth}{.80}

\renewcommand{\relwidth}{.30}

\subsubsection{Mobile multi-camera system}
\label{ss:multicamera}

Three multi-camera datasets were created from four cameras facing away from each other. In the first two datasets in this group, Mult-1 and Mult-2, only two cameras facing back-to-back were used for data acquisition.  In the Mult-1 dataset, each camera in the camera rig acquired images from only one pattern.  In Mult-2, acquisition was as for Mult-1, except that the camera rig was rotated 180 degrees such that each camera views the two patterns. Mult-3 uses all four cameras, and uses three calibration patterns. The mobile multi-camera system creates \robot \ type datasets.

\subsection{Ground truth evaluation and comparison experiment}\label{ss:comparion-ex}

For the nine Sim-\# datasets, ground truth camera poses are available.  We assess the performance of CALICO as well as Kalibr when relevant with respect to the ground truth by comparing the camera poses. First, the camera poses of both the ground truth and the results of a method are mapped to a common coordinate system, where {$^{c_0}\mathbf{C}^{\prime} = \mathbf{I}_4$}.  In other words, the first camera's HTM for the comparison $^{c_0}\mathbf{C}^{\prime}$ is the identity matrix, and all the other camera HTMs are transformed to this coordinate system by post-multiplying by the inverse of the first camera's original HTM, $^{c_i}\mathbf{C}^{\prime} = (^{c_i}\mathbf{C})\, (^{c_0}\mathbf{C}^{-1})$.

The difference in rotation angle and translation distance are then compared between the transformed camera poses of the ground truth and results from CALICO or Kalibr.  The denominator for averaging angle and distance is $\left| \mathbb{C} \right| - 1$ since there is no difference between the pose of the $c_0^{th}$ cameras between CALICO or Kalibr and the ground truth.  

We compared CALICO to the Kalibr multi-camera calibration toolbox \cite{kalibr}, using Kalibr's implementation on Github,\footnote{\href{https://github.com/ethz-asl/kalibr}{https://github.com/ethz-asl/kalibr}} and accessed this implementation through a Docker image from Stereo Labs.\footnote{\href{https://hub.docker.com/r/stereolabs/kalibr}{https://hub.docker.com/r/stereolabs/kalibr}}

CALICO compared well to the ground truth, values in Table \ref{table:syn-gt-comparison}, with a maximum rotational error of $0.23^\circ$ and translational error of $12.28$ mm.  In the case of $12.28$ mm translational error for the Sim-3-april dataset, the internal parameters for one of the cameras were estimated differently than the ground truth, leading to a different resulting camera pose. 

CALICO also compared well with Kalibr.  For the Sim-3-april dataset, as mentioned in the previous paragraph, the calibration of internal and external parameters was propagated to the constraint set $\CS$. Since for this work we chose the first camera to align between the datasets so they could be compared, and this internal parameter difference happened on the second (of two) cameras, different choices of anchor camera would likely influence results. On the Sim-4-april dataset, Kalibr's output converged but the solution was not usable; most of the twelve cameras are clustered around two cameras. For the Sim-5-april dataset, Kalibr was not able to converge. The camera rig in Sim-4-april and Sim-5-april have purely rotational motion about an axis, and from \cite{maye_online_2014, lee_2022} learned that this case is a degenerate motion for Kalibr and explain its behavior for those datasets. For the Sim-7-april and Sim-8-april datasets, Kalibr failed with the message that there was not enough mutual information. It seems that the Kalibr implementation we used requires cameras to not only image the same pattern, but also requires that cameras image a set of the same April tags.

\subsection{Results and discussion}"
\label{sec:results}

CALICO was used to calibrate the nine simulated, two stationary multi-camera, and three mobile multi-camera datasets. Quantitative results are shown in Table \ref{table:result}. Results are in terms of the three metrics: algebraic error ($ae$), reprojection root mean squared error ($rrmse$), reconstruction accuracy error ($rae$). Run time is shown as well.

Qualitatively, we found that camera poses were similar to those expected from the configuration of the multi-camera datasets. For most of the datasets, the reconstructed calibration patterns are very close to the source patterns.  

\begin{table} 
	\caption{Comparison of CALICO to ground truth and Kalibr.
	}
	\begin{centering}
		{
			{
				\begin{threeparttable}
					\begin{tabular}{c|c|c|c|c}
						\hline
						{Dataset} & \multicolumn{2}{c|}{CALICO} & \multicolumn{2}{c}{Kalibr}\\ 
						& \multrow{Rotation\\error} & \multrow{Translation\\error} &  \multrow{Rotation\\error} & \multrow{Translation\\error} \\
						\hline
						\hline
						Sim-1  & 0.234$^\circ$ & 8.565 & \multicolumn{2}{c}{did not compare} \\
						\hline
						Sim-2 &  0.062$^\circ$ & 1.368 & \multicolumn{2}{c}{did not compare} \\
						\hline
						Sim-3-april & 0.203$^\circ$ & 12.282 & 0.167$^\circ$ & 3.657 \\
						\hline
						Sim-4-april & 0.093$^\circ$ & 6.280 & 30.714$^\circ$ & 107.756 \\
						\hline
						Sim-5-april & 0.018$^\circ$ & 0.759 & \multicolumn{2}{c}{no convergence}\\
						\hline 
						Sim-6-april & 0.037$^\circ$ & 0.692 & 0.0257$^\circ$ & 0.357\\
						\hline 
					    Sim-7-april & 0.146$^\circ$ & 3.605 & \multicolumn{2}{c}{not enough mutual info}\\
						\hline 
						Sim-8-april & 0.029$^\circ$ & 2.73 & \multicolumn{2}{c}{not enough mutual info}\\
						\hline 
					    Sim-9-april & 0.022$^\circ$ & 0.563 & 0.006$^\circ$ & 0.322\\
						\hline 
					\end{tabular}
					\begin{tablenotes}
						\small {
							\item Mean rotation error (degrees) and average translation error (mm). Details in \ref{ss:comparion-ex}.  
							\item Datasets Sim-1 and Sim-2 use charuco patterns and Kalibr reads April Tags, so comparison with Kalibr is not possible with these datasets. 
						}
					\end{tablenotes}
				\end{threeparttable}
			}
		}
	\end{centering}
	\label{table:syn-gt-comparison}
\end{table}

\begin{table}
	\caption{Results of using CALICO to compute camera pose for nine datasets.}
	
	\begin{center}
		\resizebox{0.49\textwidth}{!}{
		\begin{threeparttable}
			\begin{tabular}{c|r|r|r|r|r}
				\hline
				{Dataset} &  {$ae$} &  {$rrmse$} & {$rae$} & \multicolumn{2}{c}{run time (s)} \\ 
				&  & (pixels) & (mm) & \multrow{load,\\internal\\calibration} & \multrow{multi-\\camera\\calibration} \\
				\hline
				Net-1  &  3.818  & 1.107  &  0.382 &  80.310  & 0.219 \\
				\hline
				Net-2 & 5.598  & 0.941  & 0.323 &   25.0 \quad  & 0.618   \\
				\hline
				Mult-1 & 0.960 & 0.401 & 0.607 &  2.6 \quad & 0.076 \\
				\hline
				Mult-2 & 0.862 & 0.342 & 0.34 \quad &  3.43 & 0.085 \\
				\hline 
				Mult-3 & 2.013  & 0.464  & 0.701  & 11.46 & 0.282  \\
				\hline 
				Sim-1 & 48.938 & 0.450 & 0.235 & 7.1 \quad  & 0.403 \\
				\hline 
				Sim-2 & 1.725   &  0.370   & 0.078  & 22.65 & 1.39 \quad  \\
				\hline
				Sim-3-april & 11.652  & 0.430 & 1.054 & 18.5 \quad & 0.242 \\
				\hline
				Sim-4-april & 56.592  & 0.438 & 1.102 & 202.0 \quad & 1.732 \\
				\hline
				Sim-5-april & 10.083  & 0.215 & 0.06 \quad &  194.17 & 1.572 \\
				\hline
				Sim-6-april & 0.391  & 0.386 & 0.084 &  23.3 \quad & 0.313 \\
				\hline
				Sim-7-april & 18.62 \quad & 0.438 & 0.161 &  17.0 \quad & 0.137 \\
				\hline
				Sim-8-april & 2.40 \quad & 0.41 & 4.63 \quad &  11.8 \quad & 0.096 \\
				\hline
				Sim-9-april & 1.80 \quad  & 0.41 & 0.813 &  11.6 \quad & 0.109 \\
				\hline
			\end{tabular}
		\begin{tablenotes}
			\small {
			\item Algebraic error, $ae$ (Eq. \ref{eq:algebraic_error}) has no units.
			}
		\end{tablenotes}
	\end{threeparttable}
}
	\end{center}
	\label{table:result}
\end{table}

Quantitatively, the camera poses found with CALICO had subpixel $rrmse$ values, except for dataset Net-1.  The higher $rrmse$ value of that experiment versus the others is perhaps explained by that experiment's small number of time instants (10) versus a comparably large number of cameras (12), Table \ref{table:description}.  For all of the datasets, CALICO produced a less than $1.11$ mm mean reconstruction accuracy error $rae$.

From Table \ref{table:result}, algebraic error seems not well related to the quality of results that are important to reconstruction tasks like $rae$ and $rrmse$.  While algebraic error is used to generate initial solutions, reconstruction error may be high for views of the calibration patterns where the estimation of the pattern to camera transformation ${}^{c}\mathbf{A}_{p}^{t}$ is not reliable.  The stationary multi-camera datasets have a high proportion of images in this category, so we hypothesize that this is why the algebraic error is higher for those datasets.

We also show run time in Table \ref{table:result} divided into two categories: (a) tasks preliminary to multi-camera calibration, such as image loading and individual camera calibration, and (b) multi-camera calibration. We parallelize all portions of the process. The run time is dominated by the image loading and individual camera calibration tasks. The largest datasets -- twelve cameras and 100 images each -- had low run times of 3:24 minutes for Sim-4-april and 3:15 minutes for Sim-5-april. 

\section{Conclusions}

We presented CALICO, a method for multi-camera calibration suitable for stationary multi-camera systems, mobile multi-camera systems, multi-camera systems with non-overlapping fields of view, and/or non-synchronized systems.  The method's performance was demonstrated on fourteen datasets and compared to another multi-camera calibration method, Kalibr. CALICO uses one or more calibration patterns, and so does not depend on being able to move camera rigs to a richly textured scene or and does not require synchronization. We formulated the multi-camera calibration problem within this context as a set of rigidity constraints between cameras, pattern transformations, and time labels and minimize for reprojection error. We demonstrated that the multi-camera calibration part, which excludes data loading and individual camera calibration, was computed in less than 2 seconds for all datasets. 

\section*{Acknowledgments}
A. Tabb was supported by USDA-ARS Project 8080-21000-032-00-D. T.T. Santos' work is supported by the S{\~a}o Paulo Reseach Foundation (FAPESP, grant 2017/19282-7).

\bibliographystyle{IEEEtran}
\bibliography{IEEEabrv,CalibrationGeneral}

\begin{thebibliography}{10}
\providecommand{\url}[1]{#1}
\csname url@samestyle\endcsname
\providecommand{\newblock}{\relax}
\providecommand{\bibinfo}[2]{#2}
\providecommand{\BIBentrySTDinterwordspacing}{\spaceskip=0pt\relax}
\providecommand{\BIBentryALTinterwordstretchfactor}{4}
\providecommand{\BIBentryALTinterwordspacing}{\spaceskip=\fontdimen2\font plus
\BIBentryALTinterwordstretchfactor\fontdimen3\font minus
  \fontdimen4\font\relax}
\providecommand{\BIBforeignlanguage}[2]{{%
\expandafter\ifx\csname l@#1\endcsname\relax
\typeout{** WARNING: IEEEtran.bst: No hyphenation pattern has been}%
\typeout{** loaded for the language `#1'. Using the pattern for}%
\typeout{** the default language instead.}%
\else
\language=\csname l@#1\endcsname
\fi
#2}}
\providecommand{\BIBdecl}{\relax}
\BIBdecl

\bibitem{tabb_amy_2019_3520866}
\BIBentryALTinterwordspacing
A.~Tabb and M.~J. Feldmann, ``Data and code from: Multi-camera calibration with
  pattern rigs, including for non-overlapping cameras: Calico,'' Dec. 2023.
  [Online]. Available: \url{https://zenodo.org/doi/10.5281/zenodo.3520865}
\BIBentrySTDinterwordspacing

\bibitem{zhang_flexible_2000}
\BIBentryALTinterwordspacing
Z.~Zhang, ``A flexible new technique for camera calibration,'' \emph{Pattern
  Analysis and Machine Intelligence, IEEE Transactions on}, vol.~22, no.~11,
  pp. 1330--1334, Nov. 2000. [Online]. Available:
  \url{https://doi.org/10.1109/34.888718}
\BIBentrySTDinterwordspacing

\bibitem{cook_acronym:_2019}
\BIBentryALTinterwordspacing
B.~A. Cook, ``{ACRONYM}: {Acronym} {CReatiON} for {You} and {Me},''
  \emph{arXiv:1903.12180 [astro-ph]}, Mar. 2019, arXiv: 1903.12180. [Online].
  Available: \url{http://arxiv.org/abs/1903.12180}
\BIBentrySTDinterwordspacing

\bibitem{garrido-jurado_automatic_2014}
\BIBentryALTinterwordspacing
S.~Garrido-Jurado, R.~Mu{\~n}oz{-}Salinas, F.~Madrid-Cuevas, and
  M.~Mar{\'i}n-Jim{\'e}nez, ``\BIBforeignlanguage{en}{Automatic generation and
  detection of highly reliable fiducial markers under occlusion},''
  \emph{\BIBforeignlanguage{en}{Pattern Recognition}}, vol.~47, no.~6, pp.
  2280--2292, Jun. 2014. [Online]. Available:
  \url{https://doi.org/10.1016/j.patcog.2014.01.005}
\BIBentrySTDinterwordspacing

\bibitem{tabb2019calibration}
A.~Tabb, H.~Medeiros, M.~J. Feldmann, and T.~T. Santos, ``Calibration of
  asynchronous camera networks: Calico,'' 2019.

\bibitem{shah_overview_2012}
\BIBentryALTinterwordspacing
M.~Shah, R.~D. Eastman, and T.~Hong, ``\BIBforeignlanguage{en}{An overview of
  robot-sensor calibration methods for evaluation of perception systems},'' in
  \emph{\BIBforeignlanguage{en}{Proceedings of the {Workshop} on {Performance}
  {Metrics} for {Intelligent} {Systems} - {PerMIS} '12}}.\hskip 1em plus 0.5em
  minus 0.4em\relax College Park, Maryland: ACM Press, 2012, p.~15. [Online].
  Available: \url{http://dl.acm.org/citation.cfm?doid=2393091.2393095}
\BIBentrySTDinterwordspacing

\bibitem{li_bo_multiple-camera_2013}
\BIBentryALTinterwordspacing
{Li, Bo}, L.~Heng, K.~Koser, and M.~Pollefeys, ``\BIBforeignlanguage{en}{A
  multiple-camera system calibration toolbox using a feature descriptor-based
  calibration pattern}.''\hskip 1em plus 0.5em minus 0.4em\relax IEEE, Nov.
  2013, pp. 1301--1307. [Online]. Available:
  \url{http://ieeexplore.ieee.org/document/6696517/}
\BIBentrySTDinterwordspacing

\bibitem{liu_caliber:_2016}
\BIBentryALTinterwordspacing
A.~Liu, S.~Marschner, and N.~Snavely, ``\BIBforeignlanguage{en}{Caliber:
  {Camera} {Localization} and {Calibration} {Using} {Rigidity}
  {Constraints}},'' \emph{\BIBforeignlanguage{en}{International Journal of
  Computer Vision}}, vol. 118, no.~1, pp. 1--21, May 2016. [Online]. Available:
  \url{https://doi.org/10.1007/s11263-015-0866-1}
\BIBentrySTDinterwordspacing

\bibitem{kalibr}
P.~Furgale, J.~Rehder, and R.~Siegwart, ``Unified temporal and spatial
  calibration for multi-sensor systems,'' in \emph{2013 IEEE/RSJ International
  Conference on Intelligent Robots and Systems}, 2013, pp. 1280--1286.

\bibitem{maye_self-supervised_2013}
J.~Maye, P.~Furgale, and R.~Siegwart, ``Self-supervised calibration for robotic
  systems,'' in \emph{2013 {IEEE} {Intelligent} {Vehicles} {Symposium} ({IV})},
  Jun. 2013, pp. 473--480, iSSN: 1931-0587.

\bibitem{maye_online_2014}
\BIBentryALTinterwordspacing
J.~Maye, ``\BIBforeignlanguage{en}{Online {Self}-{Calibration} for {Robotic}
  {Systems}},'' Doctoral {Thesis}, ETH Zurich, 2014, accepted:
  2017-06-13T23:36:32Z. [Online]. Available:
  \url{https://www.research-collection.ethz.ch/handle/20.500.11850/154673}
\BIBentrySTDinterwordspacing

\bibitem{olson2011tags}
E.~Olson, ``{AprilTag}: A robust and flexible visual fiducial system,'' in
  \emph{Proceedings of the {IEEE} International Conference on Robotics and
  Automation ({ICRA})}.\hskip 1em plus 0.5em minus 0.4em\relax IEEE, May 2011,
  pp. 3400--3407.

\bibitem{schonberger_structure--motion_2016}
\BIBentryALTinterwordspacing
J.~L. Sch{\"o}nberger and J.-M. Frahm,
  ``\BIBforeignlanguage{en}{Structure-from-{Motion} {Revisited}},'' in
  \emph{\BIBforeignlanguage{en}{2016 {IEEE} {Conference} on {Computer} {Vision}
  and {Pattern} {Recognition} ({CVPR})}}.\hskip 1em plus 0.5em minus
  0.4em\relax Las Vegas, NV, USA: IEEE, Jun. 2016, pp. 4104--4113. [Online].
  Available: \url{https://doi.org/10.1109/CVPR.2016.445}
\BIBentrySTDinterwordspacing

\bibitem{esquivel_calibration_2007}
\BIBentryALTinterwordspacing
S.~Esquivel, F.~Woelk, and R.~Koch, ``\BIBforeignlanguage{en}{Calibration of a
  {Multi}-camera {Rig} from {Non}-overlapping {Views}},'' in
  \emph{\BIBforeignlanguage{en}{Pattern {Recognition}}}, F.~A. Hamprecht,
  C.~Schn{\"o}rr, and B.~J{\"a}hne, Eds.\hskip 1em plus 0.5em minus 0.4em\relax
  Berlin, Heidelberg: Springer Berlin Heidelberg, 2007, vol. 4713, pp. 82--91,
  series Title: Lecture Notes in Computer Science. [Online]. Available:
  \url{http://link.springer.com/10.1007/978-3-540-74936-3_9}
\BIBentrySTDinterwordspacing

\bibitem{lebraly_calibration_2010}
\BIBentryALTinterwordspacing
P.~L{\'e}braly, E.~Royer, O.~Ait-Aider, and M.~Dhome,
  ``\BIBforeignlanguage{en}{Calibration of {Non}-{Overlapping}
  {Cameras}---{Application} to {Vision}-{Based} {Robotics}},'' in
  \emph{\BIBforeignlanguage{en}{Procedings of the {British} {Machine} {Vision}
  {Conference} 2010}}.\hskip 1em plus 0.5em minus 0.4em\relax Aberystwyth:
  British Machine Vision Association, 2010, pp. 10.1--10.12. [Online].
  Available: \url{http://www.bmva.org/bmvc/2010/conference/paper10/index.html}
\BIBentrySTDinterwordspacing

\bibitem{carrera_slam-based_2011}
\BIBentryALTinterwordspacing
G.~Carrera, A.~Angeli, and A.~J. Davison,
  ``\BIBforeignlanguage{en}{{SLAM}-based automatic extrinsic calibration of a
  multi-camera rig},'' in \emph{\BIBforeignlanguage{en}{2011 {IEEE}
  {International} {Conference} on {Robotics} and {Automation}}}.\hskip 1em plus
  0.5em minus 0.4em\relax Shanghai, China: IEEE, May 2011, pp. 2652--2659.
  [Online]. Available: \url{http://ieeexplore.ieee.org/document/5980294/}
\BIBentrySTDinterwordspacing

\bibitem{lin_infrastructure-based_2020}
Y.~Lin, V.~Larsson, M.~Geppert, Z.~Kukelova, M.~Pollefeys, and T.~Sattler,
  ``\BIBforeignlanguage{en}{Infrastructure-{Based} {Multi}-camera {Calibration}
  {Using} {Radial} {Projections}},'' in \emph{\BIBforeignlanguage{en}{Computer
  {Vision} – {ECCV} 2020}}, ser. Lecture {Notes} in {Computer} {Science},
  A.~Vedaldi, H.~Bischof, T.~Brox, and J.-M. Frahm, Eds.\hskip 1em plus 0.5em
  minus 0.4em\relax Cham: Springer International Publishing, 2020, pp.
  327--344.

\bibitem{thirthala_radial_2012}
\BIBentryALTinterwordspacing
S.~Thirthala and M.~Pollefeys, ``\BIBforeignlanguage{en}{Radial {Multi}-focal
  {Tensors}},'' \emph{\BIBforeignlanguage{en}{International Journal of Computer
  Vision}}, vol.~96, no.~2, pp. 195--211, Jan. 2012. [Online]. Available:
  \url{https://doi.org/10.1007/s11263-011-0463-x}
\BIBentrySTDinterwordspacing

\bibitem{mishra_look_2022}
\BIBentryALTinterwordspacing
S.~Mishra, S.~Nagesh, S.~Manglani, G.~Mills, P.~Chakravarty, and G.~Pandey,
  ``Look {Both} {Ways}: {Bidirectional} {Visual} {Sensing} for {Automatic}
  {Multi}-{Camera} {Registration},'' Aug. 2022, arXiv:2208.07362 [cs].
  [Online]. Available: \url{http://arxiv.org/abs/2208.07362}
\BIBentrySTDinterwordspacing

\bibitem{chen_heterogeneous_2019}
H.~Chen and S.~Schwertfeger, ``Heterogeneous {Multi}-sensor {Calibration} based
  on {Graph} {Optimization},'' in \emph{2019 {IEEE} {International}
  {Conference} on {Real}-time {Computing} and {Robotics} ({RCAR})}, Aug. 2019,
  pp. 158--163.

\bibitem{pollok_visual_2016}
T.~Pollok and E.~Monari, ``A visual {SLAM}-based approach for calibration of
  distributed camera networks,'' in \emph{2016 13th {IEEE} {International}
  {Conference} on {Advanced} {Video} and {Signal} {Based} {Surveillance}
  ({AVSS})}, Aug. 2016, pp. 429--437.

\bibitem{robinson_robust_2017}
A.~Robinson, M.~Persson, and M.~Felsberg, ``\BIBforeignlanguage{en}{Robust
  {Accurate} {Extrinsic} {Calibration} of {Static} {Non}-overlapping
  {Cameras}},'' in \emph{\BIBforeignlanguage{en}{Computer {Analysis} of
  {Images} and {Patterns}}}, ser. Lecture {Notes} in {Computer} {Science},
  M.~Felsberg, A.~Heyden, and N.~Krüger, Eds.\hskip 1em plus 0.5em minus
  0.4em\relax Cham: Springer International Publishing, 2017, pp. 342--353.

\bibitem{tabb_parameterizations_2015}
\BIBentryALTinterwordspacing
A.~Tabb and K.~M. Ahmad~Yousef, ``Parameterizations for reducing camera
  reprojection error for robot-world hand-eye calibration,'' in \emph{2015
  {IEEE}/{RSJ} {International} {Conference} on {Intelligent} {Robots} and
  {Systems} ({IROS})}, Sep. 2015, pp. 3030--3037. [Online]. Available:
  \url{http://doi.org/10.1109/IROS.2015.7353795}
\BIBentrySTDinterwordspacing

\bibitem{tabb_solving_2017}
\BIBentryALTinterwordspacing
A.~Tabb and K.~M. Ahmad Yousef, ``Solving the robot-world hand-eye(s)
  calibration problem with iterative methods,'' \emph{Machine Vision and
  Applications}, vol.~28, no.~5, pp. 569--590, Aug. 2017. [Online]. Available:
  \url{https://doi.org/10.1007/s00138-017-0841-7}
\BIBentrySTDinterwordspacing

\bibitem{triggs_bundle_2000}
\BIBentryALTinterwordspacing
B.~Triggs, P.~F. McLauchlan, R.~I. Hartley, and A.~W. Fitzgibbon,
  ``\BIBforeignlanguage{en}{Bundle {Adjustment} — {A} {Modern}
  {Synthesis}},'' in \emph{\BIBforeignlanguage{en}{Vision {Algorithms}:
  {Theory} and {Practice}}}, G.~Goos, J.~Hartmanis, J.~van Leeuwen, B.~Triggs,
  A.~Zisserman, and R.~Szeliski, Eds.\hskip 1em plus 0.5em minus 0.4em\relax
  Berlin, Heidelberg: Springer Berlin Heidelberg, 2000, vol. 1883, pp.
  298--372, series Title: Lecture Notes in Computer Science. [Online].
  Available: \url{http://link.springer.com/10.1007/3-540-44480-7_21}
\BIBentrySTDinterwordspacing

\bibitem{bradski_opencv_2000}
G.~Bradski, ``The {OpenCV} {Library},'' \emph{Dr. Dobb's Journal of Software
  Tools}, 2000.

\bibitem{shah_comparing_2011}
\BIBentryALTinterwordspacing
M.~Shah, ``\BIBforeignlanguage{en}{Comparing two sets of corresponding six
  degree of freedom data},'' \emph{\BIBforeignlanguage{en}{Computer Vision and
  Image Understanding}}, vol. 115, no.~10, pp. 1355--1362, Oct. 2011. [Online].
  Available: \url{https://doi.org/10.1016/j.cviu.2011.05.007}
\BIBentrySTDinterwordspacing

\bibitem{shah_solving_2013}
\BIBentryALTinterwordspacing
------, ``\BIBforeignlanguage{en}{Solving the {Robot}-{World}/{Hand}-{Eye}
  {Calibration} {Problem} {Using} the {Kronecker} {Product}},''
  \emph{\BIBforeignlanguage{en}{Journal of Mechanisms and Robotics}}, vol.~5,
  no.~3, p. 031007, Jun. 2013. [Online]. Available:
  \url{https://doi.org/10.1115/1.4024473}
\BIBentrySTDinterwordspacing

\bibitem{ceres-solver}
S.~Agarwal, K.~Mierle, and Others, ``Ceres solver,''
  \url{http://ceres-solver.org}.

\bibitem{hartley_triangulation_1997}
\BIBentryALTinterwordspacing
R.~I. Hartley and P.~Sturm, ``\BIBforeignlanguage{en}{Triangulation},''
  \emph{\BIBforeignlanguage{en}{Computer Vision and Image Understanding}},
  vol.~68, no.~2, pp. 146--157, Nov. 1997. [Online]. Available:
  \url{https://www.sciencedirect.com/science/article/pii/S1077314297905476}
\BIBentrySTDinterwordspacing

\bibitem{hartley_defense_1997}
\BIBentryALTinterwordspacing
R.~Hartley, ``\BIBforeignlanguage{en}{In defense of the eight-point
  algorithm},'' \emph{\BIBforeignlanguage{en}{IEEE Transactions on Pattern
  Analysis and Machine Intelligence}}, vol.~19, no.~6, pp. 580--593, Jun. 1997.
  [Online]. Available: \url{http://ieeexplore.ieee.org/document/601246/}
\BIBentrySTDinterwordspacing

\bibitem{hartley_multiple_2004}
R.~I. Hartley and A.~Zisserman, \emph{Multiple {View} {Geometry} in {Computer}
  {Vision}}, 2nd~ed.\hskip 1em plus 0.5em minus 0.4em\relax Cambridge
  University Press, ISBN: 0521540518, 2004.

\bibitem{lee_2022}
J.~Lee, D.~Hanley, and T.~Bretl, ``Extrinsic calibration of multiple inertial
  sensors from arbitrary trajectories,'' \emph{IEEE Robotics and Automation
  Letters}, vol.~7, no.~2, pp. 2055--2062, 2022.

\end{thebibliography}

\end{document}